\documentclass[manuscript,screen]{acmart}

\AtBeginDocument{%
  \providecommand\BibTeX{{%
    \normalfont B\kern-0.5em{\scshape i\kern-0.25em b}\kern-0.8em\TeX}}}

\setcopyright{acmcopyright}
\copyrightyear{2022}
\acmYear{2022}
\acmDOI{XXXXXXX.XXXXXXX}

\begin{document}

\title{When ChatGPT for Computer Vision Will Come? From 2D to 3D}

\author{Chenghao Li}
\affiliation{%
  \institution{KAIST}
  \country{South Korea}}
\email{lch17692405449@gmail.com}

\author{Chaoning Zhang}
\affiliation{%
  \institution{Kyung Hee University}
  \country{South Korea}
}
\email{chaoningzhang1990@gmail.com}

\renewcommand{\shortauthors}{Li and Zhang, et al.}

\begin{abstract}
ChatGPT and its improved variant GPT4 have revolutionized the NLP field with a single model solving almost all text related tasks. However, such a model for computer vision does not exist, especially for 3D vision. This article first provides a brief view on the progress of deep learning in text, image and 3D fields from the model perspective. Moreover, this work further discusses how AIGC evolves from the data perspective. On top of that, this work presents an outlook on the development of AIGC in 3D from the data perspective.

\end{abstract}



\maketitle
{
  \hypersetup{linkcolor=black}
  \tableofcontents
}

\vspace{20pt}

\section{Introduction}\label{sec:intro}

Generative AI (AIGC)~\cite{zhang2023complete} has made remarkable progress in the past few years, for which ChatGPT~\cite{zhang2023one} is a phenomenal product, attracting significant attention. The advent of ChatGPT is largely attributed to large-scale pre-trained models such as BERT~\cite{devlin2018bert} and GPTs~\cite{radford2018improving,radford2019language,brown2020language,openai2023gpt4}. These models have not only performed well on natural language processing tasks~\cite{daniel2007speech}, but also provided strong support for computer vision~\cite{szeliski2022computer} and other fields.

In the text field, large-scale deep learning models~\cite{han2021pre} represented by ChatGPT have made revolutionary achievements in natural language processing (NLP) tasks. They use a large amount of text data for pre-training, thereby achieving a high level of understanding and generation of natural language. These models have been widely used in tasks such as machine translation~\cite{koehn2009statistical}, text summarization~\cite{allahyari2017text}, Q\&A systems~\cite{adamopoulou2020overview}, and have shown performance beyond human in various evaluations. These successful applications provide new ideas for researchers, further driving the development of generative AI~\cite{zhang2023complete} in the text field. Meanwhile, in the image field, large-scale deep learning models have also made important breakthroughs. These models use a large amount of image data for training, providing strong support for computer vision tasks. They have achieved excellent results in image recognition~\cite{wu2015image}, object detection~\cite{zou2019object}, semantic segmentation~\cite{long2015fully} and other tasks, leading the research direction of computer vision field. Generative AI has also made remarkable breakthroughs in the image field, laying a solid foundation for the application of artificial intelligence in image processing and analysis. Compared with 2D images, the progress in 3D domain is relatively lagging~\cite{shi2022deep}. 3D models have great potential in simulating the real world, digital media, and virtual reality. While text-to-image can generate high-resolution, exquisite images, text-to-3D models are not yet able to achieve the same results. With the help of NeRF, the 3D domain can take another route, using the powerful priors from text-to-image to assist in training text-to-3D models. Moreover, some deep learning based 3D model reconstruction and generation algorithms have made great progress~\cite{mildenhall2021nerf, poole2022dreamfusion, nichol2022point}, laying the foundation for future large-scale pre-trained models in the 3D field. Overall, compared with the text and image fields, the 3D field faces more challenges in terms of data scale, computational complexity, and model representation.

The rest of this article is orgenized as follows. Section~\ref{sec:model} discusses the development of deep learning from the model perspective. Moreover, Section~\ref{sec:data} further covers the develpment of AIGC from the data perspective. On top of that, Section~\ref{sec:outlook} presents an outlook on the development of AIGC in 3D from the data perspective.

\section{Model perspective} \label{sec:model}

\begin{figure}[h]
    \centering
    \includegraphics[width=\linewidth]{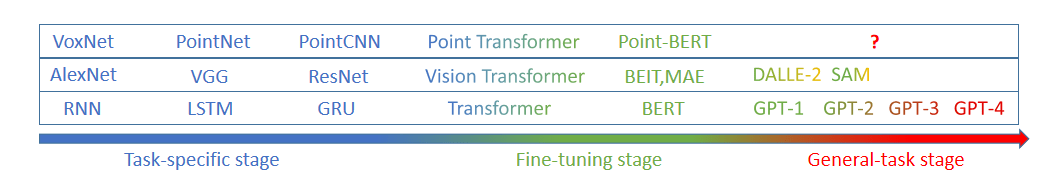}
    \caption{Three stages of deep learning in the fields of text, image and 3D, and their representative works.}
    \label{fig:3Stages}
\end{figure}

\textbf{Three stages of deep learning}. Since the success of AlexNet in 2012~\cite{krizhevsky2017imagenet}, deep learning has developed rapidly, which can be regarded as a paragon of data-driven AI. Three important factors influencing the development of deep learning are data, computing power and model. During the evolution of these three factors, deep learning can be roughly divided into three stages: \textit{task-specific stage}, \textit{fine-tuning stage} and \textit{general-task stage} by their performance, as shown in Figure~\ref{fig:3Stages}.

Firstly, in the task-specific stage, due to the limited amount of data, models with strong inductive bias, such as convolutional neural networks (CNNs)~\cite{he2016deep, szegedy2015going, tan2019efficientnet} and recurrent neural networks (RNNs)~\cite{cho2014learning}, as well as supervised learning, were adopted. These models can achieve good results on a small amount of data.

Then, in the large model pre-training and fine-tuning stage, the training data became more complex and extensive, and a small number of model parameters and network layers could no longer learn the complex and extensive data distribution. The mainstream model structure also became globally dependent transformer~\cite{vaswani2017attention}, using self-supervised learning, first learning complex features, and then fine-tuning to solve downstream tasks.

At present, text has entered the general task stage. With more data and super-large models, the distribution of data closer to reality can be learned with only specific prompts to complete free content generation. Representative works at this stage include the GPT series~\cite{radford2018improving, radford2019language, brown2020language, openai2023gpt4} in the text domain. In the image field, there is no unified CV field tasks for the large model yet, but Dalle~\cite{ramesh2022hierarchical, ramesh2021zero} has shown strong text-prompted 2d image generation capabilities, and SAM (Segment Anything)~\cite{kirillov2023segment} has demonstrated a dominant ability in the single segmentation task.

\textbf{Text.} With the development of deep learning, the Natural Language Processing (NLP) field has entered the era of super-large models. The transformer~\cite{vaswani2017attention} model introduced by Google is the underlying network architecture of the current large model, and ChatGPT released by OpenAI in November 2022 is a dialog-type super-large language model that uses the generative pre-trained transformer (GPT) to process sequence data, with language understanding and text generation capabilities. It can train models through massive corpus to achieve almost indistinguishable chat scenarios from real humans. ChatGPT can not only be used as a chatbot, but also for tasks such as writing emails, video scripts, copywriting, translation, and coding. The NLP field is in the forefront in the era of super-large models.

\textbf{Image.} On top of GPT-3, OpenAI released the large-scale image generation model DALL-E, which can generate digital images from natural language descriptions, referred to as “prompts”. Later, OpenAI released DALL-E 2, a successor designed to generate more realistic images and higher resolution, with 3.5B parameters. It can combine concepts, attributes and styles. This model has achieved considerable results in the field of computer vision, and has made significant improvements in the quality of generation. More recently, Meta has released a project termed ``Segment Anything"~\cite {kirillov2023segment}, which introduces a new task called promptable segmentation together with a new large segmentation dataset. The resulting model termed segment anything model (SAM) mimics the GPT-3 in the NLP to adopt prompt engineering for adapting it to various downstream taks, which demonstrates impressive zero-shot transfer performance. Numerous works have either evaluated its capability of generalization~\cite{tang2023can,han2023segment} or robustness~\cite{zhang2023attacksam}. Moreover, some works have combined SAM with other models to realize image editing and inpainting~\cite{GroundedSegmentAnything2023,kevmo2023magiccopy,feizc2023IEA}, which shows that SAM can also help generative tasks.

The exploration of pre-training stage is currently underway in 3D field~\cite{he2021deep, shi2022deep, yu2022point}. The research in the field of 3D deep learning is limited by computation power and data access, lagging behind image and text. This is because 3D data is more complex, scarce and diversified than images and texts. Existing 3D content creation approaches usually require a lot of professional knowledge and manpower, which can be very time-consuming and expensive. There have been numerous pioneering attempts to study how to automatically generate 3D data\cite{mildenhall2021nerf, poole2022dreamfusion, nichol2022point}. However, the quality and generalization of the generated samples are far inferior to those of large models in text and image.

\begin{figure}
    \centering
    \includegraphics[width=\linewidth]{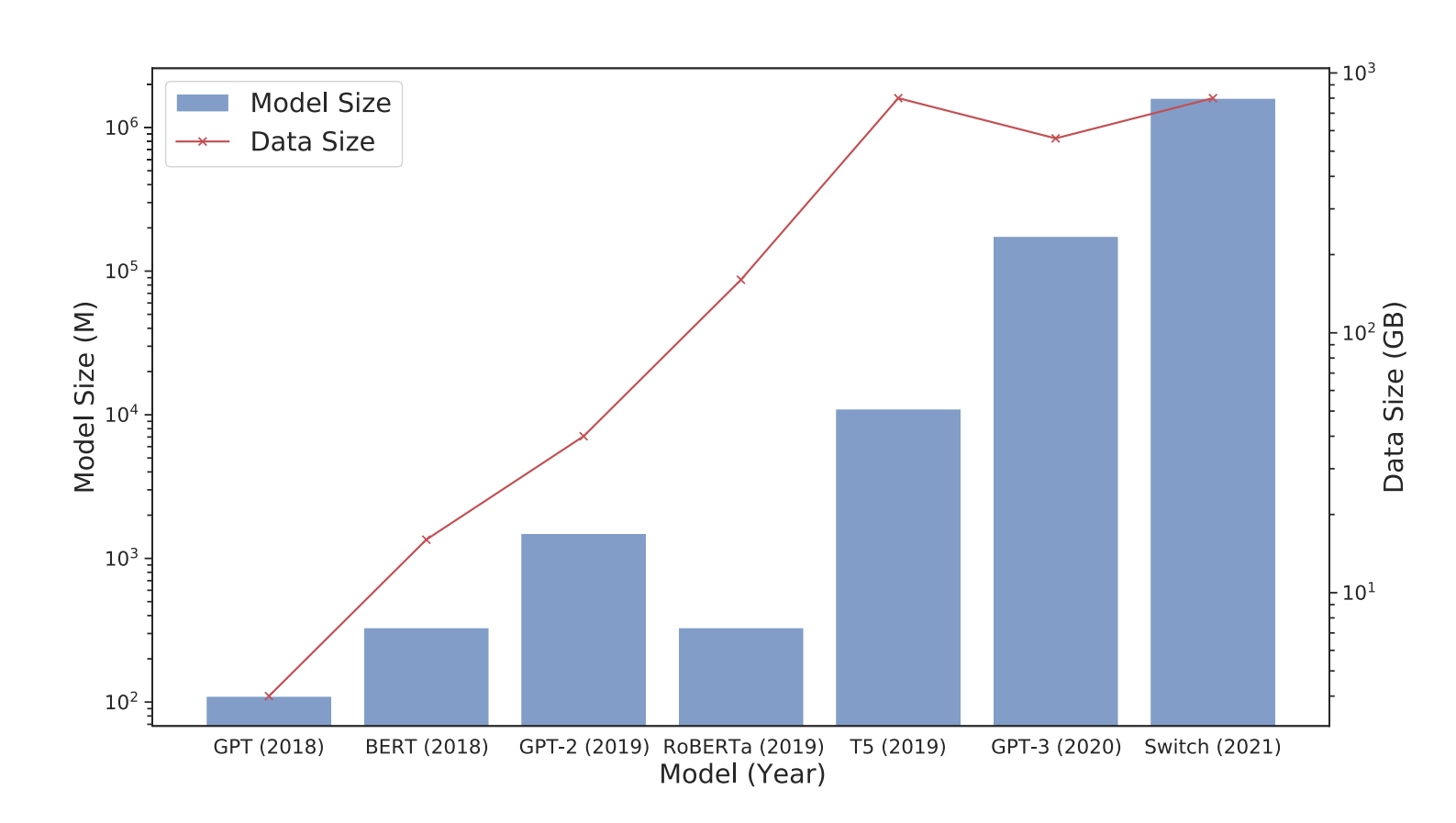}
    \caption{From the perspective of training data volume and model parameter volume, the development process of NLP large model from 2018 to 2021. The picture is derived from~\cite{han2021pre}.}
    \label{fig:NLP-Bert2GPT}
\end{figure}

\begin{figure}[h]
    \centering
    \includegraphics[width=\linewidth]{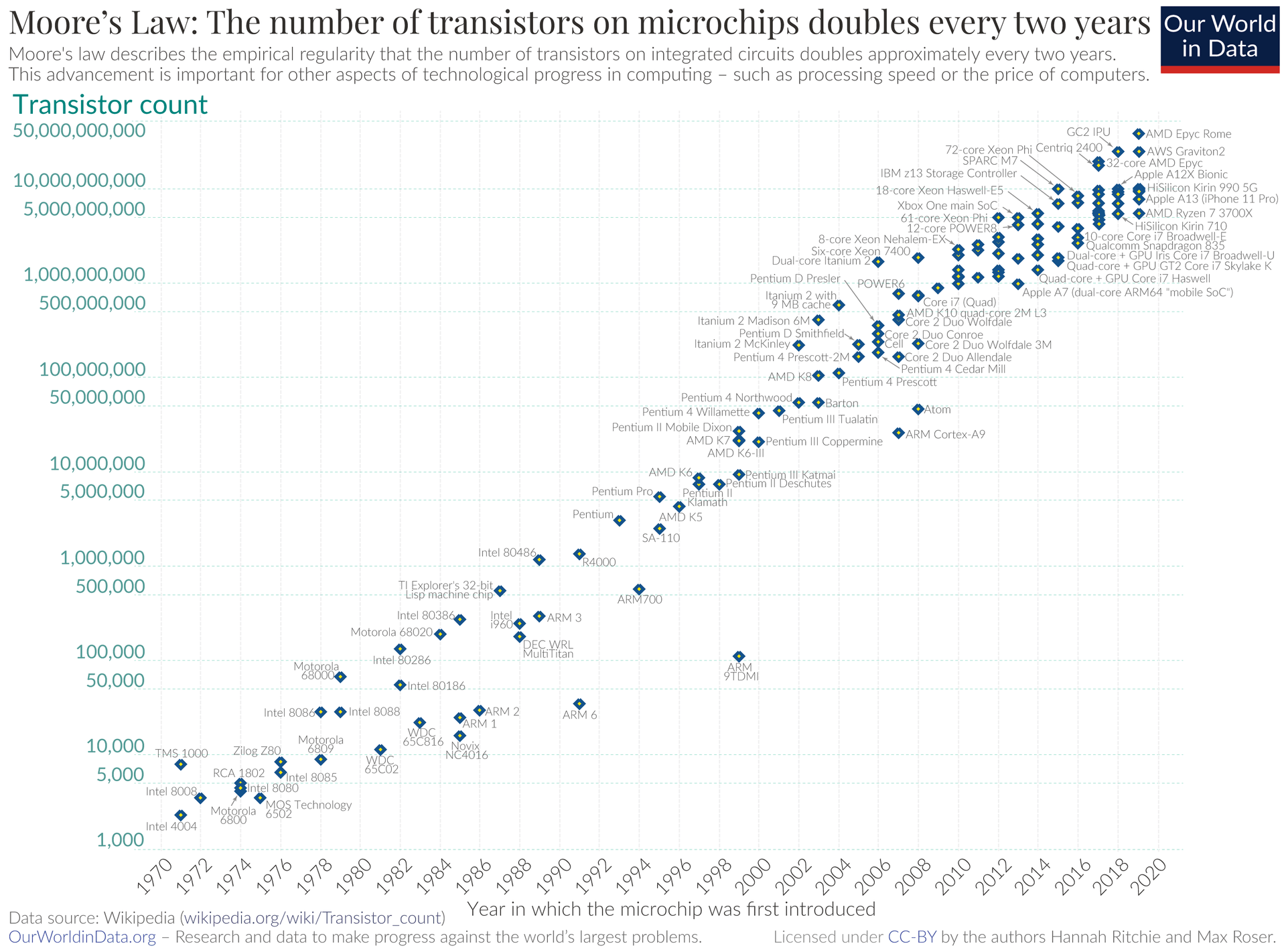}
    \caption{A logarithmic graph showing the timeline of how transistor counts in microchips are almost doubling every two years from 1970 to 2020; Moore's Law. The picture is derived from~\cite{Roser2020Moore}.}
    \label{fig:moorelaw}
\end{figure}

The development of large models (including training methods and model structures) is determined by two main factors: \textit{Computational power} and \textit{Data volume}. This also indirectly affects the development of large models, which can be referenced by the development history of large models in the NLP field, as shown in the figure \ref{fig:NLP-Bert2GPT}. The number of model parameters and the amount of training data are limited by computing power. By referring to Moore's Law (as shown in Figure \ref{fig:moorelaw}), we can assume and predict the development of computing power. So how to solve the data problem, in other words, \textit{when will the explosion of three-dimensional data come?}

\section{Data perspective} \label{sec:data}

Before we discuss the issue of data explosion in the 3D field, let us first 
look into how PGC and UGC promote AIGC. PGC (Professionally Generated Content) refers to content that is produced and published by professionals, institutions, or companies. Such content usually has a high quality, reliability, and authority, such as news reports, films, television programs, etc. UGC (User Generated Content) is content created and shared by ordinary users or community members. Compared with PGC, the quality of UGC is uneven, but it provides a platform for ordinary users to express their opinions and ideas. Typical UGC platforms include social media, blogs, forums, etc. AIGC (AI Generated Content) refers to content created by artificial intelligence algorithms and systems. AIGC covers a variety of content types, such as text, images, music, etc., and its quality is gradually improving with the progress of technology, gradually approaching or even surpassing the level of human creation. 

The relationship between PGC, UGC and AIGC is complementary. Each of PGC, UGC and AIGC has its own unique advantages and limitations. PGC has authority and reliability, but the cost of creation is relatively high and the speed of updating is relatively slow. UGC has wide range and diversity, but the quality is uneven. The development of AIGC needs time, but after the generative AI is mature, it can generate unexpected high-quality content in a short time.

\begin{figure}[h]
    \centering
    \includegraphics[width=0.85\linewidth]{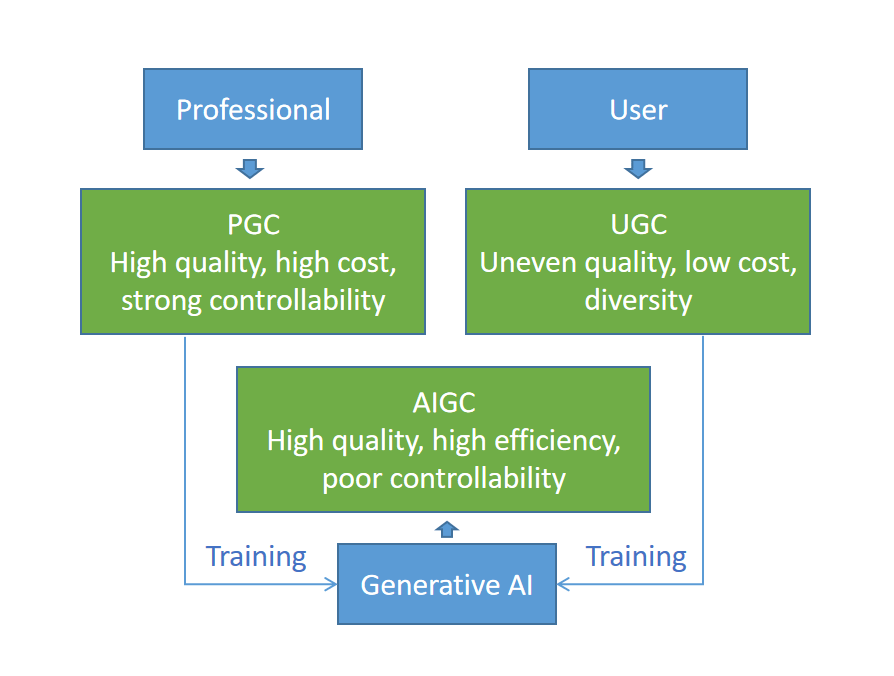}
    \caption{Three data generation methods, PGC and UGC promote AIGC by providing training data.}
    \label{fig:3GC}
\end{figure}

The emergence order of these three in text and image fields is PGC, UGC and AIGC. At the beginning of the development of the Internet, content was mainly created by professionals and institutions, i.e. PGC. With the development of Internet technology and platforms, users began to participate in content creation, forming UGC. In recent years, with the rapid development of artificial intelligence technology, AI has gradually entered content creation, forming AIGC.

The emergence of AIGC in both text and image domains is largely dependent on PGC and UGC. From a technical perspective, the development of AIGC is indeed benefited from the large amount of data from PGC and UGC. In the process of AI generating content, PGC and UGC provide AI with rich samples and knowledge, helping AI to learn different types of content and styles. In this sense, the development of AIGC depends on the existence of PGC and UGC. However, from another perspective, even without PGC and UGC, AIGC still may be developed. In theory, as long as there is enough technical progress and innovation, AI can learn and generate content independently. But in practical applications, the existence of PGC and UGC provides AI with a more rich and diverse learning environment, which accelerates the development of AI in the field of content creation. 

Overall, there is a close relationship between PGC, UGC and AIGC. The development of AIGC has benefited a lot from the existence of PGC and UGC, which provides AI with abundant learning resources. However, theoretically speaking, the development of AIGC does not entirely depend on the existence of PGC and UGC. In the future, with the continuous progress of AI technology, we believe that AIGC will play an even more important role in content creation.

\section{Outlook} \label{sec:outlook}

\subsection{Data explosion since 2010}

\begin{figure}[h]
    \centering
    \includegraphics[width=\linewidth]{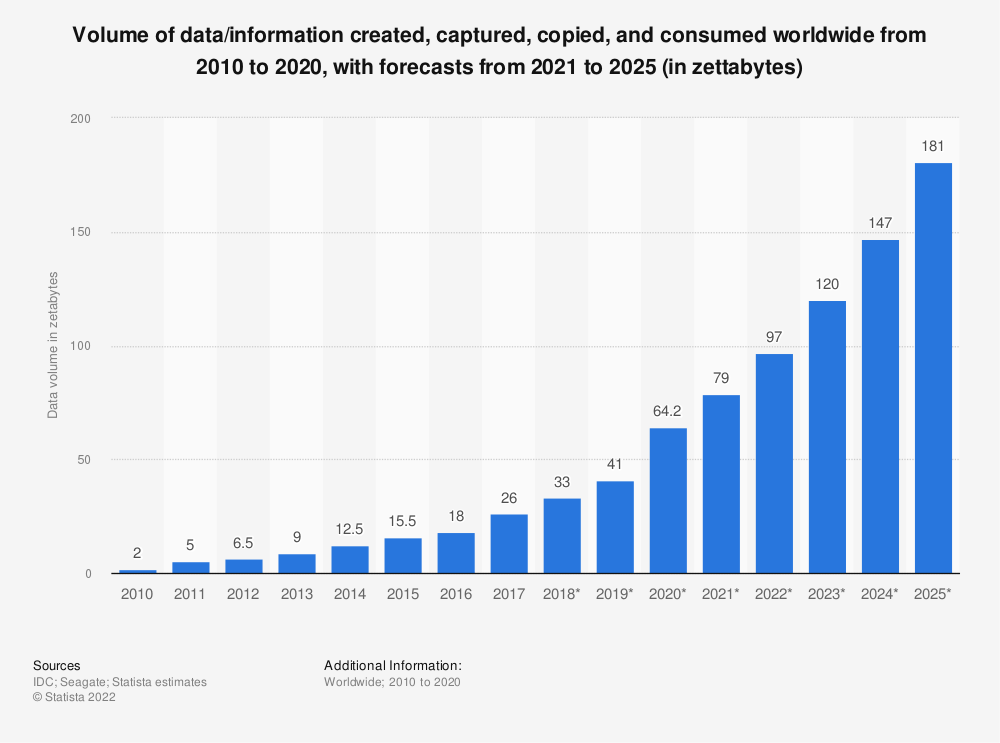}
    \caption{Amount of data created consumed and stored from 2010 to 2020 with forecasts to 2025. The picture is derived from~\cite{idc2021data}.}
    \label{fig:2010-2020}
\end{figure}

The emergence of IoT devices especially smart phone greatly increased the demand for network text and image data in several aspects. The amount of data created consumed and stored change since 2010 is visually shown in the figure~\ref{fig:2010-2020}. The rise of social media was one of the main factors. At the time of the release of iPhone 4, social media such as Facebook, Twitter and Instagram were becoming increasingly popular and the portability and features of smart phones made them ideal devices for users to communicate and share content on these platforms, resulting in a rapid growth in the demand for text and image data on social media platforms. The continuously upgraded camera technology has enabled mobile phone users to easily take high-quality photos and videos, allowing users to upload and share more image data on social media, photo sharing platforms, and other online services. 

The expansion of the App Store was also an important factor. With the development of the smartphone, the ecosystem of the App Store was further developed, with more developers creating applications for mobile, including social media applications, photo editing applications, games, news readers, etc., making it easier for users to access and share text and image data and further increasing the demand for such data. Finally, with the increasing demand for high-quality text and image data, content creators and companies began to invest more energy and resources into creating and publishing such content, leading to a rapid increase in the amount of text and image data. Overall, the emergence of smartphones, with their innovations and improvements, has greatly increased the demand for text and image data, which has to some extent contributed to the development of foundation models in those fields.

\subsection{The 3D field awaits its own 'iPhone4'}

In the field of technology, the iPhone 4 is not only a mobile phone, but also an important milestone. Its emergence made the global audience amazed and changed the pattern of the smartphone industry. Today, the 3D field is also in need of a product with the same influence to promote the development and popularization of the industry. We are looking forward to the birth of a 3D product that can lead the trend and make people rethink the value of this technology. In the past few years, 3D technology has made great progress, however, there is still a certain distance from the mainstream market acceptance and wide applications. We can think from the following aspects how the 3D field can welcome its "iPhone 4".

First of all, 3D products need to have stronger usability. Just as the iPhone 4 improved the touchscreen, operating system and user interface, making the phone more intuitive and easy to use, 3D technology also needs to make breakthroughs in this area. Both hardware and software need to provide users with an intuitive operating experience, reduce the learning cost, and make it easier for more people to get started. Secondly, innovative business models and application scenarios are crucial. The iPhone 4 created a huge application market for developers, attracting the birth of countless wonderful applications. The 3D field also needs to find a similar breakthrough, so that 3D technology is no longer limited to certain specific scenarios, but become an indispensable part of our daily life. Finally, the 3D field needs a leading enterprise to lead innovation. Apple has become the leader of the smartphone market with the success of the iPhone 4. The 3D field also needs an enterprise with vision and strength to drive technological progress and market competition. Such an enterprise can not only lead the market trend, but also drive the technological innovation and application expansion of the entire industry. Overall, in the 3D field, a popular consumer-level product is expected to promote the demand for 3D data, thus further promoting the development of deep learning in the 3D field.

\section{Conclusion}
Given that ChatGPT dominates the NLP field for solving almost all text tasks with a single model, this work provides a brief view on when such a model might come to the computer vision field, that ranges from 2D vision to 3D vision. 

\bibliographystyle{ACM-Reference-Format}
\bibliography{sample-base}

\end{document}